\documentclass[runningheads]{llncs}
\usepackage[T1]{fontenc}
\usepackage{graphicx,verbatim}

\usepackage{amsmath,amssymb}
\usepackage{amsmath, amssymb, amsfonts}
\usepackage{dsfont}

\usepackage{booktabs}
\usepackage{multirow}
\usepackage{makecell}
\usepackage{array}
\usepackage{enumitem}
\usepackage{float}
\usepackage{marvosym}

\usepackage{booktabs}   
\usepackage{multirow}   
\usepackage{colortbl}   
\usepackage{xcolor}     
\usepackage{graphicx} 
\usepackage{tabularx}
\definecolor{highlightcolor}{RGB}{236, 242, 255}

\usepackage{xcolor}
\definecolor{linkblue}{rgb}{0.21,0.49,0.74}

\usepackage[colorlinks=true,linkcolor=linkblue,citecolor=linkblue,urlcolor=linkblue,bookmarks=true,bookmarksopen=true]{hyperref}
\hypersetup{
    pdfborder={0 0 0},  
    pdfstartview={FitH}, 
}

\makeatother
\usepackage{tikz}
\usetikzlibrary{arrows.meta, positioning, shapes.geometric}

\makeatletter

\begin{document}
\title{Medical SAM3: A Foundation Model for Universal Prompt-Driven Medical Image Segmentation}
\titlerunning{Medical SAM3 for Medical Image Segmentation}
%

\author{
Chongcong Jiang\inst{1}\textsuperscript{*} \and
Tianxingjian Ding\inst{1}\textsuperscript{*} \and
Chuhan Song\inst{2}\textsuperscript{*} \and
Jiachen Tu\inst{3} \and
Ziyang Yan\inst{4}\textsuperscript{$\dagger$} \and
Yihua Shao\inst{5} \and
Zhenyi Wang\inst{1} \and
Yuzhang Shang\inst{1} \and
Tianyu Han\inst{6} \and
Yu Tian\inst{1}\textsuperscript{$\dagger$\Letter}
}

\authorrunning{Jiang et al.}
\institute{
    University of Central Florida, Orlando, USA \and
    University College London, London, UK \and
    University of Illinois Urbana-Champaign, Champaign, USA \and 
    University of Trento, Trento, Italy \and
    The Hong Kong Polytechnic University, China \and 
    University of Pennsylvania, Philadelphia, USA 
    \\
    \email{yu.tian2@ucf.edu}}

\maketitle
\begingroup
\renewcommand{\thefootnote}{}
\footnotetext{* Co-first authors. $\dagger$ Corresponding to Z. Yan and Y. Tian.}
\endgroup    
\vspace{-25pt}
\begin{abstract}
Promptable segmentation foundation models such as SAM3 have demonstrated strong generalization capabilities through interactive and concept-based prompting. However, their direct applicability to medical image segmentation remains limited by severe domain shifts, the absence of privileged spatial prompts, and the need to reason over complex anatomical and volumetric structures.
Here we present Medical SAM3, a foundation model for universal prompt-driven medical image segmentation, obtained by fully fine-tuning SAM3 on large-scale, heterogeneous 2D and 3D medical imaging datasets with paired segmentation masks and text prompts. Through a systematic analysis of vanilla SAM3, we observe that its performance degrades substantially on medical data, with its apparent competitiveness largely relying on strong geometric priors such as ground-truth-derived bounding boxes. These findings motivate full model adaptation beyond prompt engineering alone. 
By fine-tuning SAM3's model parameters on 33 datasets spanning 10 medical imaging modalities, Medical SAM3 acquires robust domain-specific representations while preserving prompt-driven flexibility. Extensive experiments across organs, imaging modalities, and dimensionalities demonstrate consistent and significant performance gains, particularly in challenging scenarios characterized by semantic ambiguity, complex morphology, and long-range 3D context.
Our results establish Medical SAM3 as a universal, text-guided segmentation foundation model for medical imaging and highlight the importance of holistic model adaptation for achieving robust prompt-driven segmentation under severe domain shift. Code and model will be made available
at \url{https://github.com/AIM-Research-Lab/Medical-SAM3}.

\keywords{Medical Image Segmentation \and Foundation Models \and Fine-Tuning \and SAM3}

\end{abstract}

\section{Introduction}
\label{sec:intro}

\begin{figure}[!t]
  \centering  \includegraphics[width=0.75\textwidth]{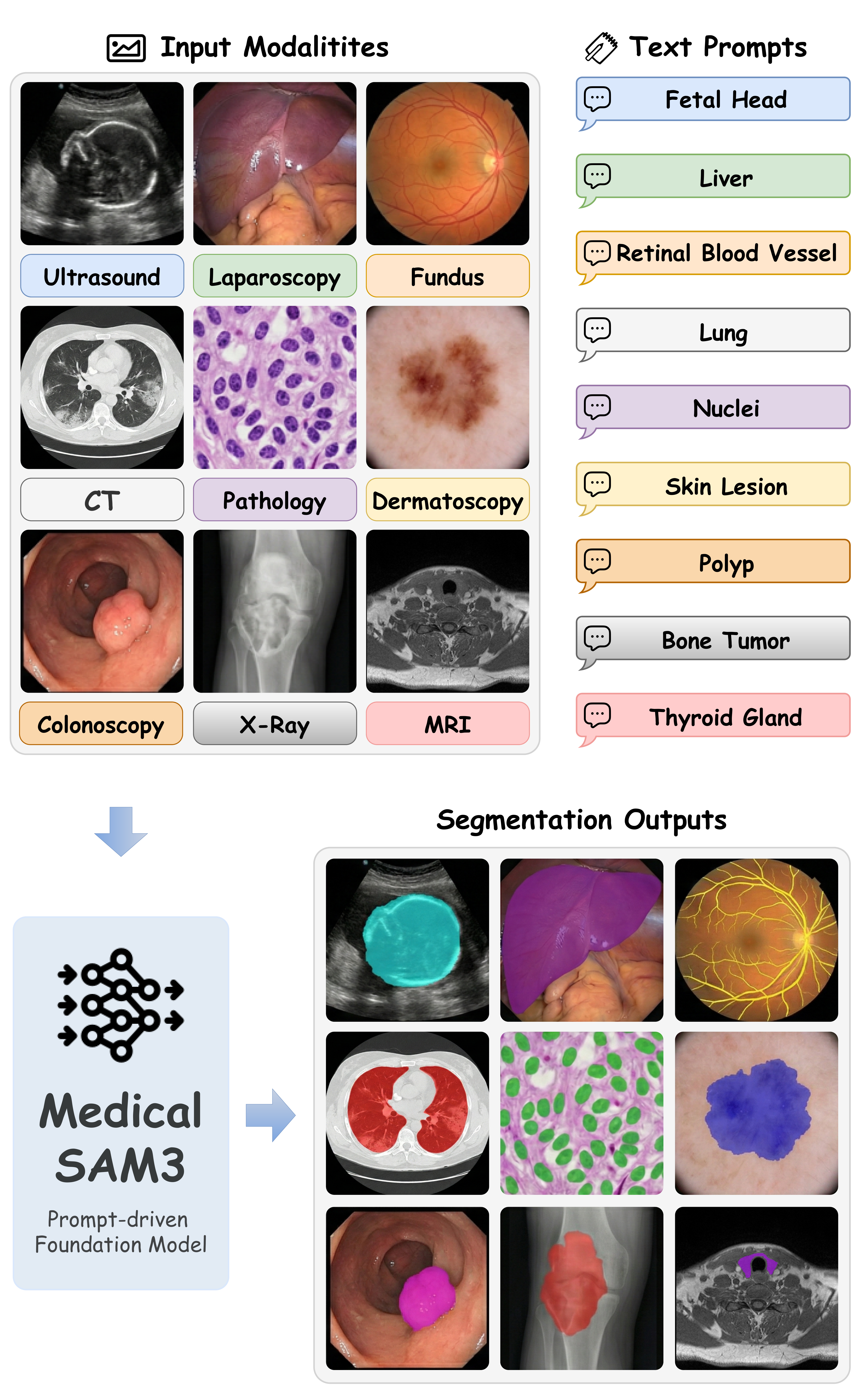} 
  \vspace{-15pt}
  \caption{\textbf{Universal medical image segmentation via text prompting with Medical SAM3.} 
  Our proposed model unifies diverse medical imaging modalities—ranging from radiology (CT, MRI, X-Ray) to optical imaging (Fundus, Dermoscopy, Endoscopy) and pathology—into a single framework. }
  \vspace{-15pt}
  \label{fig:teaser}
\end{figure}

Medical image segmentation aims to delineate clinically relevant structures and abnormalities in medical images at the pixel or voxel level. By enabling objective quantification of disease extent and anatomical changes, segmentation supports lesion assessment, surgical or radiotherapy planning, and longitudinal follow-up~\cite{litjens2017survey,esteva2017dermatologist,shen2017deep}. Despite remarkable progress in deep learning, many models remain optimized for specific tasks and data distributions, making adaptation to new modalities, anatomies, pathologies, or clinical sites challenging~\cite{guan2021domain,liu2021feddg,shao2024gwq}. This reliance on expert dense annotation and dataset-specific optimization limits scalability and hinders deployment in long-tail rare conditions and heterogeneous real-world settings, especially under distribution shift.

Methodologically, the field has been dominated by fully supervised specialist models trained with dense annotations. Convolutional neural networks (CNNs) and vision transformers (ViTs) have achieved strong performance for medical segmentation~\cite{ronneberger2015unet,milletari2016vnet,chen2021transunet,hatamizadeh2022swinunetr}, and automated pipelines further reduce manual tuning~\cite{isensee2021nnunet}. However, these advances largely remain within a dataset-centric paradigm and do not readily generalize across modalities and clinical sites, motivating promptable foundation models that provide a more unified and scalable interface for segmentation.

Segmentation foundation models offer a promising alternative, aiming to generalize across tasks through prompt-based interaction while reducing task-specific retraining~\cite{shao2025eventvad}. The Segment Anything Model (SAM)~\cite{kirillov2023sam} demonstrated remarkable zero-shot generalization in natural images via visual prompts, and subsequent models such as SAM3~\cite{carion2025sam3} extend this paradigm with concept-based prompting. However, a critical gap remains. Medical images differ substantially from natural scenes in acquisition protocols and semantic structure, often leading to unstable performance under zero-shot or lightly adapted settings~\cite{he2024accuracy,mazurowski2023segment,huang2024segment}. More crucially, many previous foundation models achieve competitive results only by relying on ground-truth-derived bounding boxes, essentially utilizing oracle localization cues~\cite{ma2024medsam,wu2025medsa,wang2023sammed3d,shao2025tr}. While effective for interactive refinement, such privileged geometric priors largely remove the localization challenge and reduce the problem to boundary refinement, which may confound comparisons when geometric priors are not available at deployment. In real deployments, boxes must be provided by a clinician or an upstream detector. Without such cues, text-only prompts often degrade sharply under severe domain misalignment\cite{zhao2025sat,lueddecke2022clipseg,shao2025icm}, motivating holistic model adaptation for robust, prompt-driven medical segmentation.

To address these challenges, we present \textbf{Medical SAM3}, a universal prompt-driven foundation model for medical image segmentation obtained by holistically adapting SAM3 on large-scale, heterogeneous 2D and 3D medical datasets with paired segmentation masks and text prompts. By moving beyond lightweight adapters and reducing reliance on pre-defined geometric cues (e.g., bounding boxes), Medical SAM3 learns robust domain-specific representations while preserving promptable flexibility under severe domain shift. We further conduct a systematic diagnostic study of vanilla SAM3 in medical settings and evaluate Medical SAM3 across both internal validation tasks and external validation tasks spanning diverse organs, modalities, and dimensionalities. Across this suite, Medical SAM3 achieves state-of-the-art performance and supports a spatial-prompt-free, semantic-driven paradigm for medical image segmentation. In summary, our contributions are threefold: \textbf{(i)} we introduce Medical SAM3 by holistically adapting SAM3 for universal, text-guided medical segmentation without privileged spatial prompts; \textbf{(ii)} we provide a diagnostic study that characterizes the failure modes of vanilla SAM3 under severe domain shift and its reliance on geometric cues; and \textbf{(iii)} we curate a large-scale text–image–mask aligned medical segmentation corpus and establish strong results through extensive internal and external evaluations across diverse organs, modalities, and 2D/3D settings.

\section{Related Works}

\renewcommand\paragraph{\@startsection{paragraph}{4}{\z@}%
  {-12\p@ \@plus -4\p@ \@minus -4\p@}%
  {-0.5em \@plus -0.22em \@minus -0.1em}%
  {\normalfont\normalsize\bfseries}}

\paragraph{Specialist Medical Image Segmentation.}
Fully supervised specialist models remain the dominant paradigm in medical image segmentation. Early encoder--decoder CNNs and their variants, represented by FCN and U-Net, establish strong inductive biases for dense prediction and are widely extended with attention and redesigned skip connections \cite{tian2024fairdomain,long2015fcn,ronneberger2015unet,oktay2018attentionunet,zhou2018unetpp,tian2023self,tian2021constrained,tian2023unsupervised}. For volumetric imaging, 3D architectures directly model spatial context in CT and MRI, including 3D U-Net and V-Net \cite{cicek2016unet3d,milletari2016vnet}. Beyond architecture design, automated training pipelines such as nnU-Net substantially reduce manual engineering and provide strong baselines across datasets \cite{isensee2021nnunet}. Large scale multi-organ segmentation systems further demonstrate that broad anatomical coverage can be achieved when sufficient annotations and standardized pipelines are available \cite{wasserthal2023totalsegmentator,shao2025accidentblip}. More recently, Transformer based designs improve global context modeling for medical segmentation, including hybrid and fully Transformer architectures \cite{chen2021transunet,hatamizadeh2022unetr,hatamizadeh2022swinunetr,yan2025renderworld,zhou2021nnformer,tian2023fairseg}. In parallel, selective state space models (SSMs), exemplified by Mamba, have been explored to capture long range dependencies with improved efficiency, inspiring Mamba-based medical segmentation architectures such as U-Mamba, SegMamba, VM-UNet, and Swin-UMamba \cite{gu2023mamba,ma2024umamba,xing2024segmamba,ruan2024vmunet,Liu_SwinUMamba_MICCAI2024}.

\paragraph{Text Guided and Open Vocabulary Segmentation.}
Text guided segmentation in general vision is commonly approached by aligning dense visual features with language representations to enable open vocabulary mask prediction \cite{lueddecke2022clipseg,rao2022denseclip,yan20243dsceneeditor,ding2023maskclip,liang2023maskadaptedclip}. Referring expression segmentation further studies phrase grounded masks through explicit cross modal fusion \cite{shao2025context,wang2022cris,yang2022lavt}. These lines of work provide complementary perspectives on semantic conditioning and prompt design that are relevant to text based target specification in medical segmentation.

\paragraph{Promptable Segmentation Foundation Models.}
Interactive medical segmentation predates recent foundation models and commonly improves an automatic prediction with lightweight user inputs, such as clicks or scribbles, as exemplified by DeepIGeoS \cite{wang2018deepigeos}. SAM introduces a promptable interface via a prompt encoder and a mask decoder, enabling segmentation conditioned on spatial prompts \cite{kirillov2023sam}, and SAM 2 extends this design with memory for streaming image and video settings \cite{ravi2024sam2}. Medical adaptations of SAM style models have been studied through supervised domain adaptation and parameter efficient customization, including MedSAM and Medical SAM Adapter \cite{ma2024medsam,wu2025medsa}. Extending promptable segmentation to volumetric data has been explored from different perspectives, including learning native 3D promptable models such as SAM-Med3D and using memory mechanisms for 3D image/video settings such as MedSAM2 \cite{wang2023sammed3d,ma2025medsam2}. In addition, universal medical segmentation has been investigated via prompt driven multi task learning and minimal interaction paradigms, including UniSeg, MedUniSeg, and One-Prompt Segmentation \cite{ye2023uniseg,ye2024meduniseg,wu2024oneprompt}. Most recently, concept based prompting has been introduced in SAM3 to broaden conditioning beyond purely geometric cues \cite{carion2025sam3}, and large vocabulary medical segmentation driven by text prompts has also been explored \cite{zhao2025sat}.

\section{Method}
\label{sec:methodology}

\begin{figure}[t]
\centering
\makebox[\textwidth][c]{\includegraphics[width=1\textwidth]{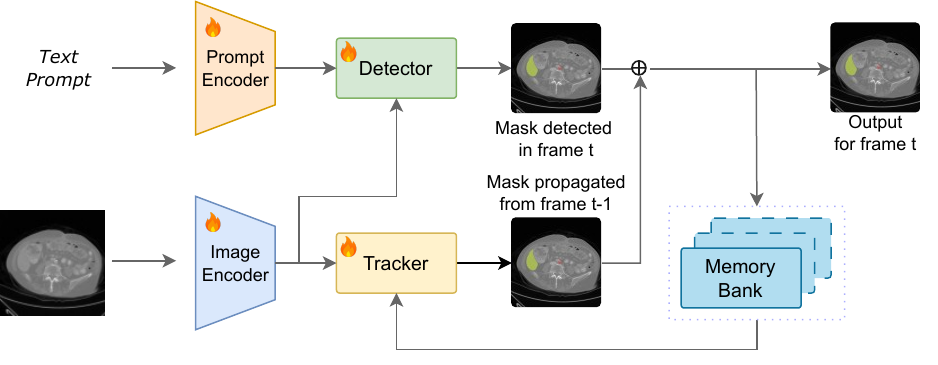}}
\caption{\textbf{Overview of Medical SAM3.} Medical SAM3 takes a text prompt and medical images (2D or slice-based 3D) as input. A detector segments target instances in the current frame, while an optional tracker propagates masks across frames via a memory bank. The final prediction is produced by merging detected and propagated masks, supporting semantic-driven segmentation without privileged spatial prompts.}

\label{fig:method_overview}
\end{figure}

In this paper, we propose a full fine-tuning strategy to adapt SAM3~\cite{carion2025sam3}, a large-scale promptable segmentation foundation model, to medical imaging under severe domain shift. Unlike parameter-efficient or partial fine-tuning approaches, we update all model parameters to enable comprehensive domain adaptation. Crucially, we introduce \textit{no architectural modifications} to SAM3. Figure~\ref{fig:method_overview} illustrates our training pipeline for 2D and 3D modalities, which follows SAM3's detector--tracker design for sequential inputs. At frame $t$, Medical SAM3 combines a mask detected by the detector with a mask propagated from slice $t{-}1$ by the tracker, and updates the memory bank for subsequent propagation.

\subsection{Unified Input Formulation}
\label{sec:unification}

Medical imaging spans a wide spectrum of departments, encompassing natively planar modalities such as Histopathology, Fundus photography, Dermatology, and Projection Radiography (X-ray). To harmonize these heterogeneous data sources into a generalist foundation model, we unify these modalities within a common 2D feature space. By treating each medical scan as a high-fidelity 2D image, we maximize the model's applicability across diverse clinical workflows without being constrained by inconsistent 3D acquisition geometries. This strategy not only simplifies the integration of diverse clinical workflows but also enables the perception backbone to prioritize high-resolution spatial features (at 1008 × 1008 pixels), which are often compromised in computationally heavy volumetric frameworks.

To leverage the 33 diverse datasets during joint training, we structure each sample into a text-driven triplet $(I, M, t)$, where $I$ is the image, $M$ is the corresponding mask, and $t$ is the text prompt derived directly from the dataset's clinical labels. Unlike traditional segmentation models that require a fixed, closed-set label space, our approach exploits the semantic flexibility of the pre-trained text encoder. By associating masks with their native clinical nomenclature, the model learns to associate varied terminology with their corresponding visual features. This strategy avoids the need for complex label re-mapping while allowing the model to internalize a vast range of anatomical and pathological descriptors across disparate medical domains.

\subsection{Stratified Tuning}
\label{sec:highres_tuning}

Medical images contain critical diagnostic details that necessitate high spatial resolution. We maintain a training resolution of $1008 \times 1008$ pixels to align with the high-frequency spatial priors inherited from the original large-scale pre-training. This ensures that the positional embeddings remain synchronized with the perception backbone.

To mitigate the significant domain gap between natural and medical textures without catastrophic forgetting, we employ Layer-wise Learning Rate Decay (LLRD). For a base learning rate $\eta_{base}$, the learning rate $\eta_l$ for the $l$-th layer of the vision backbone is defined as:
\begin{equation}
    \eta_l = \eta_{base} \cdot \gamma^{L-l}
\end{equation}
where $L=12$ is the total number of layers and $\gamma=0.85$ is the decay factor. This stratified strategy allows shallow layers to retain general-purpose visual primitives, such as edges and textures, while forcing deeper layers to specialize in complex medical semantics.

\subsection{Text-Driven Semantic Alignment}
\label{sec:text_driven_alignment}

In practical clinical environments, the requirement for manual bounding boxes as spatial priors often creates a bottleneck, as it assumes the clinician has already identified the target's precise location. To maximize the utility of Medical SAM3 as an autonomous assistant, we transition from a prompt-dependent paradigm to a strictly text-driven semantic alignment strategy. By utilizing clinical concepts as the sole input during training, we force the model to develop an intrinsic spatial awareness that bridges abstract medical nomenclature with pixel-level morphological features.

This alignment process is formulated as a semantic-to-spatial distillation task. Without the crutch of a bounding box, the transformer decoder must learn to treat the text embedding $z_{txt} = E_{txt}(c)$ not merely as a class label, but as a discriminative spatial query. Through this pure text-driven supervision, the model is compelled to identify long-range correlations between high-level clinical descriptors (e.g., ``irregular mass,'' ``calcified node'') and specialized pathological textures within the vision backbone's feature maps. This global-to-local reasoning path ensures that the linguistic manifold and the visual manifold are explicitly aligned. Consequently, at inference time, the model can interpret conceptual keywords and autonomously perform zero-shot localization, effectively simulating a clinician's cognitive process of translating a diagnostic term into a visual search.

\subsection{Set-Prediction Objective}
\label{sec:loss}

\label{sec:objective}
We optimize the model using a multi-task objective that jointly supervises instance discovery and semantic segmentation. Given predicted queries $\{\hat{\mathbf{y}}_i\}_{i=1}^{N}$ and ground-truth instances $\{\mathbf{y}_j\}_{j=1}^{M}$, we establish a one-to-one assignment $\pi$ via bipartite Hungarian matching. To address potential sparse supervision in medical scenes, an auxiliary one-to-many (O2M) matcher $\pi^{\text{o2m}}$ is employed to enhance training stability. The total objective is:
\begin{equation}
    \mathcal{L}_{\text{total}} = \mathcal{L}_{\text{find}}(\pi) + \lambda_{\text{o2m}}\mathcal{L}_{\text{find}}(\pi^{\text{o2m}}) + \mathcal{L}_{\text{seg}},
    \label{eq:total_loss}
\end{equation}
where all terms are normalized by the batch-wise matched instance count.

\paragraph{Finding Loss.} For matched queries, $\mathcal{L}_{\text{find}}$ supervises classification, presence, and localization:
\begin{equation}
    \mathcal{L}_{\text{find}}(\pi) = \lambda_{\text{ce}}\mathcal{L}_{\text{ce}} + \lambda_{\text{pr}}\mathcal{L}_{\text{pres}} + \mathds{1}_{\{j \neq \varnothing\}} (\lambda_{\ell_1}\mathcal{L}_{\ell_1} + \lambda_{\text{g}}\mathcal{L}_{\text{giou}}),
\end{equation}
where $\mathcal{L}_{\text{ce}}$ is a focal-style classification loss and $\mathcal{L}_{\text{pres}}$ supervises query presence. The box regression terms ($\ell_1$ and GIoU) are computed only for positive assignments ($j \neq \varnothing$).

\paragraph{Segmentation Loss.} To ensure precise mask boundaries—critical for clinical quantification—the segmentation loss $\mathcal{L}_{\text{seg}}$ combines pixel-wise and structural terms:
\begin{equation}
    \mathcal{L}_{\text{seg}} = \lambda_{\text{f}}\mathcal{L}_{\text{seg}}^{\text{focal}} + \lambda_{\text{d}}\mathcal{L}_{\text{dice}} + \lambda_{\text{sp}}\mathcal{L}_{\text{seg-pres}},
\end{equation}
where $\mathcal{L}_{\text{dice}}$ improves boundary adherence and $\mathcal{L}_{\text{seg-pres}}$ provides semantic presence supervision.

\section{Experiments}
\label{sec:experiments}

\subsection{Datasets}
\label{sec:datasets}
To develop a prompt-driven foundation model with strong generalization to medical segmentation tasks, we fine-tune Medical SAM3 on a diverse multi-domain collection assembled from publicly accessible datasets, where each sample is paired with a segmentation mask and a text prompt that is manually curated or derived from dataset labels. 

As shown in Table~\ref{tab:datasets}, the collected corpus encompasses 33 datasets across 10 imaging modalities—including radiography (CXR and X-ray/angiography), ultrasound, endoscopy, pathology, fundus, dermoscopy, microscopy, virtual microscopy, electron microscopy, and others—amounting to a total of 76,956 images and 263,705 mask annotations. Radiography is the largest contributor with 40,160 images, dominated by large-scale CXR collections. Ultrasound and endoscopy/fetoscopy form two mid-sized groups with 12,179 and 12,887 images, respectively, while the remaining modalities provide long-tail diversity that improves coverage of appearance, acquisition, and annotation styles. The median annotation area varies by several orders of magnitude, ranging from 58 px in electron microscopy nuclei to over one million pixels in chest radiographs, highlighting substantial scale variation in segmentation targets. For consistency, we standardize all datasets to an 85/15 split for training and validation with a fixed seed of 42, yielding approximately 65.4k training images and 11.5k validation images.

\begin{table}[t]
\centering
\caption{Summary of datasets used for training. The table reports the number of images and annotations across distinct medical modalities, and the median annotation area represents the typical scale of the segmentation targets.}
\label{tab:datasets}

\small
\setlength{\tabcolsep}{8pt}
\renewcommand{\arraystretch}{1.1}

\begin{tabular}{lrrr} 
\toprule
\textbf{Dataset} & \textbf{Images} & \textbf{Anns} & \textbf{Median Area (px)} \\
\midrule

\multicolumn{4}{l}{\textbf{\textit{CXR (3 datasets)}}} \\ 
COVID-QU-Ex              & 33,920 & 67,839 & 7,542 \\
Chest Xray Masks and Labels & 704   & 1,415  & 1,022,508 \\
Chest X-Ray Pneumothorax & 290    & 370    & 7,183 \\
\midrule

\multicolumn{4}{l}{\textbf{\textit{X-ray (2 datasets)}}} \\
BTXRD  & 3,746 & 2,273 & 14,746 \\
ARCADE & 1,500 & 2,316 & 1,472 \\
\midrule

\multicolumn{4}{l}{\textbf{\textit{Ultrasound (7 datasets)}}} \\
BUSI              & 647   & 647   & 17,348 \\
BUS-UCLM          & 264   & 281   & 24,102 \\
US-Nerve          & 2,323 & 2,323 & 6,954 \\
ACOUSLIC          & 300   & 300   & 58,330 \\
ps-fh-aop-2023    & 4,000 & 3,999 & 8,048 \\
CT2USforKidneySeg & 4,586 & 4,601 & 11,807 \\
SegThy            & 59    & 130   & 2,118 \\
\midrule

\multicolumn{4}{l}{\textbf{\textit{Endoscopy (6 datasets)}}} \\
CholecSeg8k       & 8,080 & 112,521 & 749 \\
m2caiSeg          & 614   & 804     & 102,493 \\
PolypGen          & 1,710 & 2,003   & 70,698 \\
BKAI-IGH NeoPolyp & 1,000 & 1,117   & 39,138 \\
Kvasir-SEG        & 1,000 & 1,060   & 34,489 \\
\midrule

\multicolumn{4}{l}{\textbf{\textit{Pathology (4 datasets)}}} \\
PanNuke                    & 2,540 & 21,978 & 811 \\
MonuSeg                    & 82    & 2,887  & 172 \\
Breast Cancer Segmentation & 151   & 3,495  & 15,524 \\
GlaS@MICCAI'2015           & 165   & 1,538  & 10,904 \\
\midrule

\multicolumn{4}{l}{\textbf{\textit{Virtual Microscope (2 datasets)}}} \\
MUCIC Colon Tissue          & 60  & 12,396 & 1,688 \\
MUCIC HL60 Granulocytes      & 240 & 3,987  & 4,857 \\
\midrule

\multicolumn{4}{l}{\textbf{\textit{Electron Microscopy (2 datasets)}}} \\
NucMM-Z & 62 & 581 & 58 \\
UroCell & 5  & 163 & 241 \\
\midrule

\multicolumn{4}{l}{\textbf{\textit{Microscopy (1 dataset)}}} \\
PCMMD & 3,517 & 3,519 & 26,227 \\
\midrule

\multicolumn{4}{l}{\textbf{\textit{Fundus/OCT (5 datasets)}}} \\
Intraretinal Cystoid Fluid & 1,459 & 4,601 & 202 \\
PAPILA                     & 488   & 488   & 172,486 \\
COph100                    & 324   & 324   & 7,168 \\
RAVIR Dataset              & 23    & 141   & 3,638 \\
DRIVE                      & 20    & 20    & 28,176 \\
\midrule

\multicolumn{4}{l}{\textbf{\textit{Dermoscopy \& Others (2 datasets)}}} \\
ISIC\_2018 & 2,594 & 2,594 & 429,033 \\
FetoPlac  & 483   & 994   & 6,456 \\

\midrule 
\textbf{Total} & \textbf{76,956} & \textbf{263,705} & -- \\
\bottomrule
\end{tabular}
\end{table}

Our evaluation protocol is designed to rigorously test robustness under domain shift, we evaluate Medical SAM3 on 10 internal validation tasks derived from held-out splits of the fine-tuning corpus. Complementing this, we conduct external validation on 7 segmentation tasks that were entirely excluded from the model development pipeline.

Unifying these diverse benchmarks within a prompt-driven framework requires structuring each sample as an (image, mask, text) triplet. While our architecture natively supports prompts of varying granularity, spanning from broad categories to detailed descriptive attributes, we prioritize atomic clinical concepts in this study to establish a consistent baseline. We define a dataset-specific label taxonomy and map it to a unified vocabulary of canonical concept names. Single-class datasets are assigned a single global concept; Multi-class datasets use a one to one mapping from label indices to anatomy or pathology terms defined by the dataset specification. This label-to-text dictionary ensures that each segmentation mask is paired with a consistent and standardized prompt across datasets. 

\subsection{Experimental Settings}
\label{sec:exp_setup}
For both training and evaluation, we use text-only prompts for all internal and external tasks. Prompts are instantiated by applying the label-to-text mapping protocol in Sec.~\ref{sec:datasets}, resulting in a single canonical concept term per class for both training and evaluation.
We compare the results of the original SAM3~\cite{carion2025sam3} and our Medical SAM3. The original SAM3 is evaluated using the official checkpoint without any additional training on medical data. Medical SAM3 is initialized from the same SAM3 checkpoint and obtained by full parameter fine tuning on the training splits described in Sec.~\ref{sec:datasets}.

All training and testing are implemented in PyTorch with distributed data parallelism using the NCCL backend. Experiments are conducted on one node with four NVIDIA H100 GPUs with 80GB memory. We train for up to 10 epochs and select the final checkpoint based on internal validation performance. We optimize with AdamW using $\beta_1=0.9$ and $\beta_2=0.999$. We use group-wise learning rates of $3\times 10^{-4}$ for the decoder, segmentation head, and dot-product scoring, $5\times 10^{-5}$ for the vision backbone, $5\times 10^{-5}$ for the language backbone, and $1\times 10^{-4}$ for the geometry prompt encoder. The learning rate schedule uses linear warmup followed by an inverse-square-root decay. 
Training uses only text prompts paired with ground-truth segmentation masks, without any spatial or interactive prompts such as points or bounding boxes. Model selection is based on performance on the internal validation set. We follow the set-prediction objective in Sec.~\ref{sec:loss} for instance discovery and mask prediction. We use focal Hungarian matching with an auxiliary one-to-many branch to improve assignment stability under sparse supervision and severe foreground--background imbalance. All matching and loss hyperparameters are summarized in Table~\ref{tab:loss_hparams}.

During evaluation, we select the highest-confidence mask generated from the text prompt. For multi-class scenarios, we query each class independently and resolve overlaps via pixel-wise maximal confidence, yielding a single non-overlapping semantic map. This strategy ensures consistent predictions suited for text-only deployment. We report Dice coefficient and Intersection-over-Union (IoU) as primary metrics. 


\begin{table}[H]
\vspace{-20pt}
\centering
\caption{Matching and loss hyperparameters used in the set-prediction objective.}
\label{tab:loss_hparams}

\scriptsize
\setlength{\tabcolsep}{6pt} 
\renewcommand{\arraystretch}{1.10}

\resizebox{\columnwidth}{!}{%
\begin{tabular}{@{}p{0.30\columnwidth} l r l r@{}}
\toprule
\textbf{Block} & \textbf{Param} & \textbf{Val.} & \textbf{Param} & \textbf{Val.} \\
\midrule

\multirow{3}{*}{\shortstack[l]{O2O matcher\\\texttt{BinaryHungarianMatcherV2}}}
 & $w_{\text{cls}}$ & 2.0 & $w_{\text{box}}$ & 5.0 \\
 & $w_{\text{giou}}$ & 2.0 & $\alpha_{\text{match}}$ & 0.25 \\
 & $\gamma_{\text{match}}$ & 2 & stable & \texttt{false} \\
\midrule

\multirow{2}{*}{\shortstack[l]{O2M matcher\\\texttt{BinaryOneToManyMatcher}}}
 & top-$k$ & 4 & threshold & 0.4 \\
 & $\alpha_{\text{o2m}}$ & 0.3 & $\lambda_{\text{o2m}}$ & 2.0 \\
\midrule

\multirow{4}{*}{\shortstack[l]{$\mathcal{L}_{\text{find}}$}}
 & $\lambda_{\text{ce}}$ & 20.0 & $\lambda_{\text{pr}}$ & 20.0 \\
 & $\alpha_{\text{cls}}$ & 0.25 & $\gamma_{\text{cls}}$ & 2 \\
 & pos.\ weight & 10 & padded $N_q$ & 200 \\
 & $\lambda_{\ell_1}$ & 5.0 & $\lambda_{\text{g}}$ & 2.0 \\
\midrule

\multirow{3}{*}{\shortstack[l]{$\mathcal{L}_{\text{seg}}$}}
 & $\alpha_{\text{seg}}$ & 0.6 & $\gamma_{\text{seg}}$ & 2.0 \\
 & $\lambda_{\text{f}}$ & 20.0 & $\lambda_{\text{d}}$ & 30.0 \\
 & $\lambda_{\text{sp}}$ & 1.0 &  &  \\
\bottomrule
\end{tabular}%
}
\end{table}

\section{Results}
\label{sec:results}

\begin{figure}[H]
\vspace{-20pt}
\centering
\includegraphics[width=0.8\textwidth]{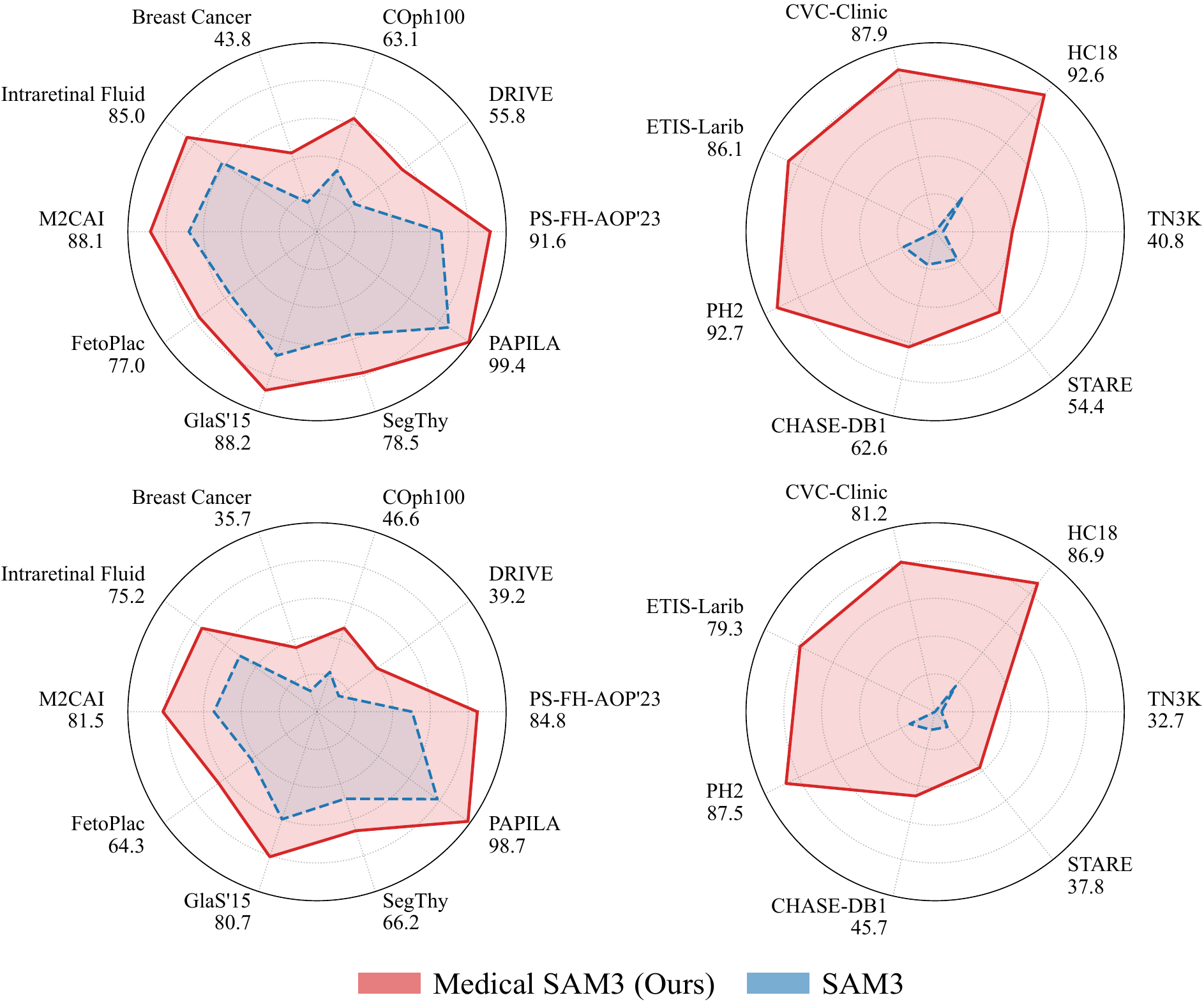}
\caption{Radar chart overview of segmentation performance. Results are split by internal validation (\textbf{top}) and external generalization (\textbf{bottom}), reporting Dice (\textbf{left}) and IoU (\textbf{right}) scores. The red area (Medical SAM3) significantly covers the blue area (SAM3) in all scenarios, aligning with the metrics in Table~\ref{tab:inter_exter_valid}.}
\end{figure}

\paragraph{Internal validation on held-out splits.}
Table~\ref{tab:inter_exter_valid} (top) reports results on 10 internal held-out splits. Medical SAM3 improves over the original SAM3 on all tasks, increasing the average Dice from 54.0\% to 77.0\% and the average IoU from 43.3\% to 67.3\%.
These gains highlight that full-parameter fine-tuning strengthens medical domain visual priors and improves text-to-mask alignment, enabling reliable localization even when only a class name is provided.
The improvements are most pronounced for small, thin, or low-contrast targets where text-only prompting is particularly challenging.
For retinal vessel segmentation, performance increases substantially on DRIVE from 24.8\% to 55.8\% Dice and on COph100 from 34.1\% to 63.1\% Dice, indicating better boundary adherence for fine vascular structures.
We also observe strong gains on modality-specific targets with large appearance shifts, including fetal head segmentation on PS-FH-AOP'23 from 65.7\% to 91.6 Dice and placental vessel segmentation on FetoPlac from 56.6\% to 77.0\% Dice. Overall, the consistent gains across all internal held-out splits indicate that Medical SAM3 achieves strong in-domain adaptation under a text-only prompting setting. 

\begin{table}[H]
\centering
\vspace{-10pt}
\caption{Quantitative comparison on internal (10) and external (7) testing datasets. We report Dice and IoU (\%).}
\label{tab:inter_exter_valid}
\footnotesize
\setlength{\tabcolsep}{4pt}
\renewcommand{\arraystretch}{1.04}

\begin{tabular}{l c >{\columncolor{highlightcolor}}c c c >{\columncolor{highlightcolor}}c c}
\toprule
\textbf{Dataset} & \multicolumn{3}{c}{\textbf{Dice (\%)}} & \multicolumn{3}{c}{\textbf{IoU (\%)}} \\
\cmidrule(lr){2-4}\cmidrule(lr){5-7}
 & \textbf{SAM3} & \textbf{Ours} & $\Delta$
 & \textbf{SAM3} & \textbf{Ours} & $\Delta$ \\
\midrule

\multicolumn{7}{l}{\textbf{Internal datasets (10)}} \\
PS-FH-AOP'23        & 65.7 & \textbf{91.6} & +25.9 & 50.3 & \textbf{84.8} & +34.5 \\
DRIVE               & 24.8 & \textbf{55.8} & +31.0 & 14.2 & \textbf{39.2} & +25.0 \\
COph100             & 34.1 & \textbf{63.1} & +29.0 & 22.1 & \textbf{46.6} & +24.6 \\
Breast Cancer       & 16.3 & \textbf{43.8} & +27.5 & 11.6 & \textbf{35.7} & +24.0 \\
Intraretinal Fluid  & 62.0 & \textbf{85.0} & +23.1 & 50.4 & \textbf{75.2} & +24.8 \\
M2CAI               & 67.7 & \textbf{88.1} & +20.4 & 54.5 & \textbf{81.5} & +27.0 \\
FetoPlac            & 56.6 & \textbf{77.0} & +20.5 & 42.9 & \textbf{64.3} & +21.4 \\
GlaS'15             & 68.9 & \textbf{88.2} & +19.4 & 59.8 & \textbf{80.7} & +21.0 \\
SegThy              & 57.3 & \textbf{78.5} & +21.2 & 48.4 & \textbf{66.2} & +17.8 \\
PAPILA              & 86.2 & \textbf{99.4} & +13.1 & 78.7 & \textbf{98.7} & +20.1 \\
\cmidrule(lr){1-7}
\textbf{Avg. (Internal)} & \textbf{54.0} & \textbf{77.0} & \textbf{+23.0} & \textbf{43.3} & \textbf{67.3} & \textbf{+24.0} \\
\midrule

\multicolumn{7}{l}{\textbf{External datasets (7)}} \\
TN3K  & 4.2  & \textbf{40.8} & +36.6 & 3.4  & \textbf{32.7} & +29.3 \\
HC18  & 23.9 & \textbf{92.6} & +68.7 & 17.3 & \textbf{86.9} & +69.6 \\
CVC   & 0.0  & \textbf{87.9} & +87.9 & 0.0  & \textbf{81.2} & +81.2 \\
ETIS  & 0.0  & \textbf{86.1} & +86.1 & 0.0  & \textbf{79.3} & +79.3 \\
PH2   & 18.4 & \textbf{92.7} & +74.3 & 14.9 & \textbf{87.5} & +72.6 \\
CHASE & 17.9 & \textbf{62.6} & +44.7 & 9.8  & \textbf{45.7} & +35.9 \\
STARE & 18.6 & \textbf{54.4} & +35.8 & 10.3 & \textbf{37.8} & +27.5 \\
\cmidrule(lr){1-7}
\textbf{Avg. (External)} & \textbf{11.9} & \textbf{73.9} & \textbf{+62.0} & \textbf{8.0} & \textbf{64.4} & \textbf{+56.4} \\
\bottomrule
\end{tabular}
\vspace{-10pt}

\end{table}
In digital pathology, Medical SAM3 markedly improves breast cancer tissue segmentation from 16.3\% to 43.8\% Dice and gland segmentation on GlaS'15 from 68.9\% to 88.2\% Dice, showing robust adaptation to stain and texture variations under the same protocol.
On high-contrast targets where the baseline is already strong, Medical SAM3 maintains or further boosts accuracy, with PAPILA reaching 99.4\% Dice and 98.7\% IoU.

\paragraph{External validation under domain shift.}
To assess zero-shot generalization, we evaluate Medical SAM3 on seven external datasets that are excluded from training, spanning ultrasound, endoscopy, and fundus photography: TN3K, HC18, CHASE\_DB1, STARE, CVC-Clinic, ETIS-Larib, and PH2.
Table~\ref{tab:inter_exter_valid} (bottom) shows consistent improvements over the original SAM3 across all tasks, with average Dice increasing from 11.9\% to 73.9\% and average IoU rising from 8.0\% to 64.4\%.
The most striking recovery occurs in endoscopic polyp segmentation (CVC and ETIS), where the baseline SAM3 suffers catastrophic failure due to weak text-visual alignment; in contrast, Medical SAM3 successfully grounds the target, achieving 87.9\% and 86.1\% Dice, respectively.
Similarly, in ultrasound (HC18) and dermatology (PH2) tasks, the model overcomes domain gaps to boost performance by over 68\%, proving its capability to reliably localize anatomical structures in unseen domains without additional adaptation.

\begin{figure}[H]
\vspace{-16pt}
\centering
\includegraphics[width=1\textwidth, trim=0 0 0 0.3cm, clip]{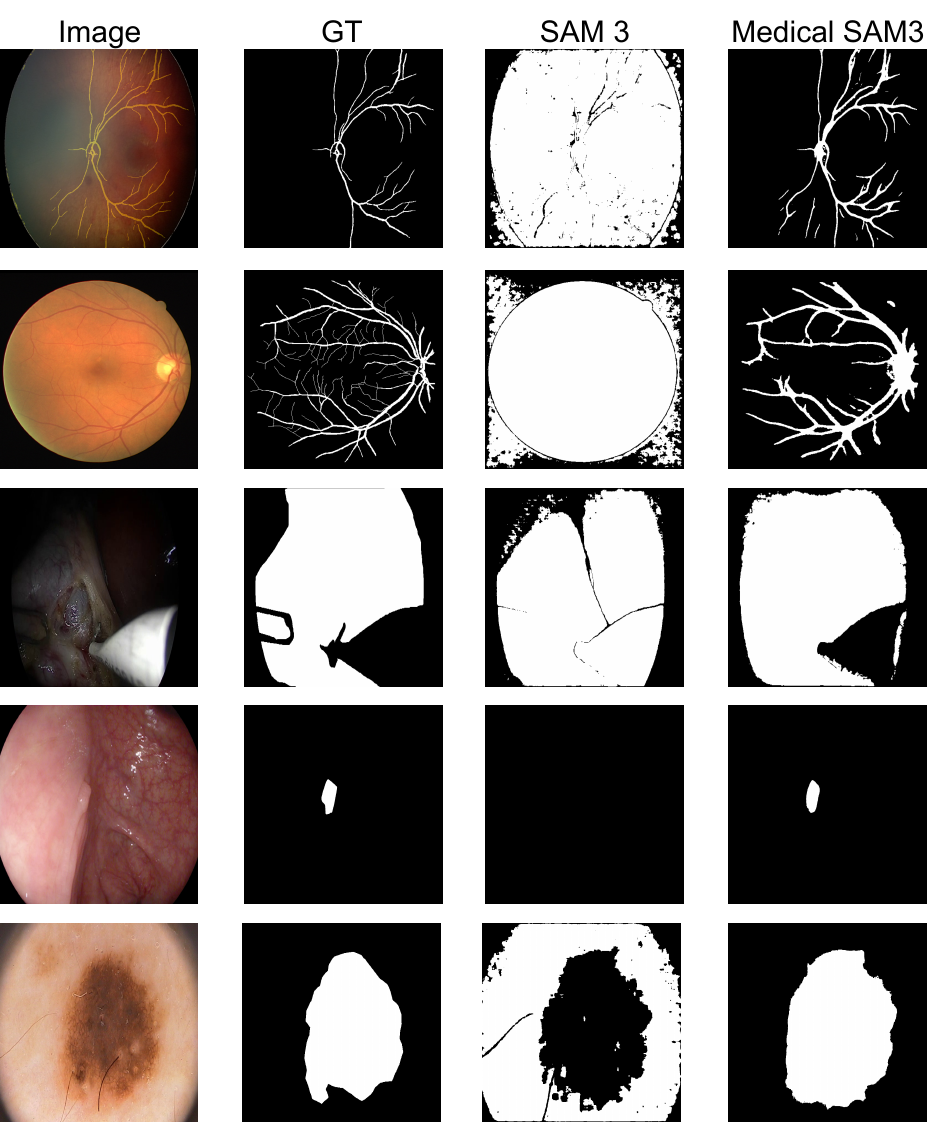}
\caption{Visualization of the segmentation performance of SAM3 and Medical SAM3}
\label{result:vis}
\vspace{-16pt}
\end{figure}

\paragraph{Qualitative Results}
Figure~\ref{result:vis} provides representative visual comparisons between SAM3 and Medical SAM3 under the same text-only prompting protocol. Across diverse modalities, the original SAM3 frequently fails to localize the target anatomy, producing either near-empty masks or severe over-segmentation that collapses to large foreground regions. This behavior is particularly evident for thin and low-contrast structures such as retinal vessels, where SAM3 outputs noisy masks with widespread false positives, while Medical SAM3 recovers fine vascular branches with substantially cleaner boundaries. Similar improvements are observed on endoscopic targets, where SAM3 tends to miss small regions or yields fragmented predictions, whereas Medical SAM3 produces coherent masks that better match the ground truth. On dermoscopy, Medical SAM3 also delineates lesion extent more accurately and avoids the spurious background activations seen in SAM3.


\section{Discussion and Conclusion}
\label{sec:discussion}

Our study reveals that the main bottleneck for universal medical segmentation with promptable foundation models is not the availability of a prompt interface, but the reliability of semantic grounding under domain shift. While strong geometric cues in medical imaging can often simplify segmentation into a boundary refinement task, relying solely on text prompts exposes the more critical challenge: mapping clinical concepts to spatially precise masks across heterogeneous appearances.

The consistent performance gains of Medical SAM3 indicate that robust text grounding is attainable when adaptation is treated as a holistic representation problem rather than merely a prompt-engineering problem. In particular, the improvements observed across diverse modalities point to a shared latent structure that can be learned when the model is forced to align high-level language concepts with localization-relevant visual features. This perspective also helps explain why failures are most visible on small, thin, or low-contrast targets: such cases demand stronger coupling between semantics and spatial evidence, and are less forgiving to misalignment.

These findings have substantial implications for both evaluation and deployment. Benchmarking should explicitly distinguish interactive settings (where users or upstream detectors provide spatial hints) from deployment-consistent semantic-only settings; otherwise, comparisons may be confounded by privileged localization priors. From a systems standpoint, a text-driven interface is attractive precisely because it offers a unified way to query segmentation targets across departments and modalities. However, realizing this promise requires standardized prompt protocols and careful handling of terminology and label granularity, since clinical language is inherently variable.

Despite these advancements, several limitations persist. First, full adaptation at high resolution can be computationally demanding, motivating future work on parameter-efficient strategies and distillation without sacrificing robustness. Second, while a planar representation improves universality across inconsistent acquisition geometries, it may underutilize native volumetric continuity; native 3D prompting and explicit inter-slice consistency constraints are promising directions. Third, our current evaluation prioritizes atomic concept prompts; extending to synonym-robust, attribute-rich, and compositional prompts will be important for real clinical usage. Finally, broader multi-center validation and reliability analyses, such as uncertainty estimation, are necessary to quantify deployment readiness.

Overall, Medical SAM3 supports a semantic-driven paradigm for universal medical segmentation and highlights that robust promptability in medicine is primarily an alignment and adaptation challenge. Future progress will likely come from combining scalable multi-domain training, richer clinical language handling, and efficiency-oriented adaptation to enable practical and trustworthy deployment.


%
\bibliographystyle{splncs04}
\bibliography{main}

@String(PAMI = {IEEE Trans. Pattern Anal. Mach. Intell.})

@String(CVPR= {IEEE Conf. Comput. Vis. Pattern Recog.})

@String(ICCV= {Int. Conf. Comput. Vis.})

@String(PAMI  = {IEEE TPAMI})

@String(CVPR  = {CVPR})

@String(ICCV  = {ICCV})

@article{tahir2021covidquex,
  title={{COVID-QU-Ex} dataset: A Large-scale Collection of COVID-19, Non-COVID, and Lung Opacity Chest X-Ray Images},
  author={Tahir, Anas M. and Chowdhury, Muhammad E. H. and Khandakar, Amith and Rahman, Tawsifur and Qiblawey, Yazan and Khurshid, Uzair and Kiranyaz, Serkan and Ibtehaz, Nabil and Rahman, M. Sohel and Al-Maadeed, Somaya and others},
  journal={Informatics in Medicine Unlocked},
  volume={30},
  pages={100893},
  year={2022},
  publisher={Elsevier}
}

@article{candemir2014lung,
  title={Lung segmentation in chest radiographs using anatomical atlases with nonrigid registration},
  author={Candemir, Sema and Jaeger, Stefan and Palaniappan, Kannappan and Musco, Jonathan P and Singh, Rahul K and Xue, Zhiyun and Karargyris, Alexandros and Antani, Sameer and Thoma, George and McDonald, Clement J},
  journal={IEEE Transactions on Medical Imaging},
  volume={33},
  number={2},
  pages={577--590},
  year={2014},
  publisher={IEEE}
}

@inproceedings{siim2019pneumothorax,
  title={{SIIM-ACR} Pneumothorax Segmentation Challenge},
  author={{Society for Imaging Informatics in Medicine}},
  booktitle={Kaggle Competition},
  year={2019},
  howpublished={\url{https://www.kaggle.com/c/siim-acr-pneumothorax-segmentation}}
}

@article{btxrd2025,
  title={{BTXRD}: Bone Tumor X-Ray Dataset for Detection and Segmentation},
  author={Wang, Xiaoming and others},
  journal={Scientific Data},
  year={2025},
  note={Bone tumor X-ray dataset}
}

@article{popescu2022arcade,
  title={{ARCADE}: Automatic Region-based Coronary Artery Disease Diagnostics Using X-Ray Angiography Images},
  author={Popescu, Dan and Diaconu, Ana-Maria and Deac, Andreea and Stanciu, Otilia and Dogaru, Radu and Bacila, C{\u{a}}lin},
  journal={Medical Image Analysis},
  volume={83},
  pages={102636},
  year={2023},
  publisher={Elsevier}
}

@article{aldhabyani2020busi,
  title={Dataset of breast ultrasound images},
  author={Al-Dhabyani, Walid and Gomaa, Mohammed and Khaled, Hussien and Fahmy, Aly},
  journal={Data in Brief},
  volume={28},
  pages={104863},
  year={2020},
  publisher={Elsevier}
}

@article{gomez2020busuclm,
  title={{BUS-UCLM}: A Breast Ultrasound Dataset for Lesion Detection and Classification},
  author={G{\'o}mez-Flores, Wilfrido and Cervantes-S{\'a}nchez, Fernando and Escalante-Ram{\'\i}rez, Boris},
  journal={Pattern Recognition Letters},
  volume={155},
  pages={33--40},
  year={2022},
  publisher={Elsevier}
}

@misc{usnerve2016,
  title={Ultrasound Nerve Segmentation Challenge},
  author={{Kaggle}},
  howpublished={\url{https://www.kaggle.com/c/ultrasound-nerve-segmentation}},
  year={2016},
  note={Brachial plexus segmentation in ultrasound images}
}

@article{acouslic2024,
  title={{ACOUSLIC-AI}: Abdominal Circumference Ultrasound Image Dataset},
  author={{ACOUSLIC Consortium}},
  journal={Grand Challenge},
  year={2024},
  howpublished={\url{https://acouslic-ai.grand-challenge.org/}}
}

@article{heuvel2018psfh,
  title={Automated measurement of fetal head circumference using 2D ultrasound images},
  author={van den Heuvel, Thomas L. A. and de Bruijn, Dagmar and de Korte, Chris L. and van Ginneken, Bram},
  journal={PloS One},
  volume={13},
  number={8},
  pages={e0200412},
  year={2018},
  publisher={Public Library of Science}
}

@article{song2022ct2us,
  title={{CT2US}: Cross-modal supervision for abdominal organ segmentation},
  author={Song, Shuangxin and others},
  journal={Medical Image Analysis},
  volume={82},
  pages={102603},
  year={2022},
  publisher={Elsevier}
}

@article{tian2023self,
  title={Self-supervised pseudo multi-class pre-training for unsupervised anomaly detection and segmentation in medical images},
  author={Tian, Yu and Liu, Fengbei and Pang, Guansong and Chen, Yuanhong and Liu, Yuyuan and Verjans, Johan W and Singh, Rajvinder and Carneiro, Gustavo},
  journal={Medical image analysis},
  volume={90},
  pages={102930},
  year={2023},
  publisher={Elsevier}
}

@inproceedings{tian2023unsupervised,
  title={Unsupervised anomaly detection in medical images with a memory-augmented multi-level cross-attentional masked autoencoder},
  author={Tian, Yu and Pang, Guansong and Liu, Yuyuan and Wang, Chong and Chen, Yuanhong and Liu, Fengbei and Singh, Rajvinder and Verjans, Johan W and Wang, Mengyu and Carneiro, Gustavo},
  booktitle={International workshop on machine learning in medical imaging},
  pages={11--21},
  year={2023},
  organization={Springer}
}

@inproceedings{tian2021constrained,
  title={Constrained contrastive distribution learning for unsupervised anomaly detection and localisation in medical images},
  author={Tian, Yu and Pang, Guansong and Liu, Fengbei and Chen, Yuanhong and Shin, Seon Ho and Verjans, Johan W and Singh, Rajvinder and Carneiro, Gustavo},
  booktitle={International Conference on Medical Image Computing and Computer-Assisted Intervention},
  pages={128--140},
  year={2021},
  organization={Springer}
}

@article{liu2024translation,
  title={Translation consistent semi-supervised segmentation for 3d medical images},
  author={Liu, Yuyuan and Tian, Yu and Wang, Chong and Chen, Yuanhong and Liu, Fengbei and Belagiannis, Vasileios and Carneiro, Gustavo},
  journal={IEEE Transactions on Medical Imaging},
  year={2024},
  publisher={IEEE}
}

@article{tian2023fairseg,
  title={Fairseg: A large-scale medical image segmentation dataset for fairness learning using segment anything model with fair error-bound scaling},
  author={Tian, Yu and Shi, Min and Luo, Yan and Kouhana, Ava and Elze, Tobias and Wang, Mengyu},
  journal={arXiv preprint arXiv:2311.02189},
  year={2023}
}

@inproceedings{tian2024fairdomain,
  title={Fairdomain: Achieving fairness in cross-domain medical image segmentation and classification},
  author={Tian, Yu and Wen, Congcong and Shi, Min and Afzal, Muhammad Muneeb and Huang, Hao and Khan, Muhammad Osama and Luo, Yan and Fang, Yi and Wang, Mengyu},
  booktitle={European Conference on Computer Vision},
  pages={251--271},
  year={2024},
  organization={Springer}
}

@article{wunderling2017segthy,
  title={{SegThy}: A Novel Dataset for Thyroid Segmentation in Ultrasound Images},
  author={Wunderling, Thomas and Golla, Bianca and Poudel, Prashnna and Taber, Clemens and Gockel, Michael and Modersitzki, Jan},
  journal={International Journal of Computer Assisted Radiology and Surgery},
  volume={12},
  number={8},
  pages={1405--1414},
  year={2017},
  publisher={Springer}
}

@inproceedings{hong2020cholecseg8k,
  title={{CholecSeg8k}: A Semantic Segmentation Dataset for Laparoscopic Cholecystectomy Based on {CholecT50}},
  author={Hong, Wang-chan and others},
  booktitle={arXiv preprint arXiv:2012.12453},
  year={2020}
}

@article{maier2017m2caiseg,
  title={Can masses of non-experts train highly accurate image classifiers? A crowdsourcing approach to instrument segmentation in laparoscopic images},
  author={Maier-Hein, Lena and others},
  journal={Medical Image Computing and Computer-Assisted Intervention (MICCAI)},
  year={2014},
  publisher={Springer}
}

@article{ali2023polypgen,
  title={{PolypGen}: A Multi-center Polyp Detection and Segmentation Dataset for Generalisability Assessment},
  author={Ali, Sharib and others},
  journal={Scientific Data},
  volume={10},
  number={1},
  pages={75},
  year={2023},
  publisher={Nature Publishing Group}
}

@article{ngoc2021bkaineopolyp,
  title={{BKAI-IGH NeoPolyp}: A Colonoscopy Dataset for Colorectal Polyp Detection and Segmentation},
  author={Ngoc Lan, Phan and others},
  journal={IEEE Access},
  volume={9},
  pages={163026--163039},
  year={2021},
  publisher={IEEE}
}

@inproceedings{jha2020kvasirseg,
  title={{Kvasir-SEG}: A Segmented Polyp Dataset},
  author={Jha, Debesh and Smedsrud, Pia H and Riegler, Michael A and Halvorsen, P{\aa}l and de Lange, Thomas and Johansen, Dag and Johansen, H{\aa}vard D},
  booktitle={International Conference on Multimedia Modeling},
  pages={451--462},
  year={2020},
  organization={Springer}
}

@article{gamper2020pannuke,
  title={{PanNuke}: An Open Pan-Cancer Histology Dataset for Nuclei Instance Segmentation and Classification},
  author={Gamper, Jevgenij and Koohbanani, Navid Alemi and Benet, Ksenija and Khuram, Ali and Rajpoot, Nasir},
  journal={European Congress on Digital Pathology},
  pages={11--19},
  year={2019},
  organization={Springer}
}

@article{kumar2019monuseg,
  title={A Dataset and a Technique for Generalized Nuclear Segmentation for Computational Pathology},
  author={Kumar, Neeraj and Verma, Ruchika and Sharma, Sanuj and Bhargava, Surabhi and Vahadane, Abhishek and Sethi, Amit},
  journal={IEEE Transactions on Medical Imaging},
  volume={36},
  number={7},
  pages={1550--1560},
  year={2017},
  publisher={IEEE}
}

@article{araujo2017breastcancer,
  title={Classification of breast cancer histology images using convolutional neural networks},
  author={Ara{\'u}jo, Teresa and Aresta, Guilherme and Castro, Eduardo and Rouco, Jos{\'e} and Aguiar, Paulo and Eloy, Catarina and Pol{\'o}nia, Ant{\'o}nio and Campilho, Aur{\'e}lio},
  journal={PloS one},
  volume={12},
  number={6},
  pages={e0177544},
  year={2017},
  publisher={Public Library of Science}
}

@article{sirinukunwattana2017glas,
  title={Gland Segmentation in Colon Histology Images: The {GlaS} Challenge Contest},
  author={Sirinukunwattana, Korsuk and Pluim, Josien PW and Chen, Hao and Qi, Xiaojuan and Heng, Pheng-Ann and Guo, Yun Bo and Wang, Ling Yang and Matuszewski, Bogdan J and Brber, Elia and others},
  journal={Medical Image Analysis},
  volume={35},
  pages={489--502},
  year={2017},
  publisher={Elsevier}
}

@article{sirinukunwattana2016mucic,
  title={Locality Sensitive Deep Learning for Detection and Classification of Nuclei in Routine Colon Cancer Histology Images},
  author={Sirinukunwattana, Korsuk and Raza, Shan E Ahmed and Tsang, Yee-Wah and Snead, David RJ and Cree, Ian A and Rajpoot, Nasir M},
  journal={IEEE Transactions on Medical Imaging},
  volume={35},
  number={5},
  pages={1196--1206},
  year={2016},
  publisher={IEEE}
}

@article{kainz2015muci,
  title={You Should Use Regression to Detect Cells},
  author={Kainz, Philipp and Urschler, Martin and Schulter, Samuel and Wohlhart, Paul and Lepetit, Vincent},
  journal={Medical Image Computing and Computer-Assisted Intervention (MICCAI)},
  pages={276--283},
  year={2015},
  publisher={Springer}
}

@article{pcmmd2025,
  title={{PCMMD}: Plasma Cell Multiple Myeloma Dataset for Cell Detection and Segmentation},
  author={{PCMMD Consortium}},
  journal={Scientific Data},
  year={2025},
  note={Microscopy dataset for plasma cell segmentation}
}

@article{lin2021nucmmz,
  title={{NucMM-Z}: A dataset for nuclear segmentation in zebrafish brain using electron microscopy},
  author={Lin, Zudi and Wei, Donglai and Liao, Jinglin and Xu, Xiaolong and Bhagat, Sanjna and others},
  journal={MICCAI},
  year={2021},
  publisher={Springer}
}

@article{levinshtein2020urocell,
  title={{UroCell}: A Dataset for Electron Microscopy Segmentation of Urinary Bladder Cells},
  author={Levinshtein, Alex and others},
  journal={Data in Brief},
  volume={30},
  pages={105522},
  year={2020},
  publisher={Elsevier}
}

@article{kovalyk2022papila,
  title={{PAPILA}: Dataset with fundus images and clinical data of both eyes of the same patient for glaucoma assessment},
  author={Kovalyk, Oleksandr and Morales-S{\'a}nchez, Juan and Verd{\'u}-Monedero, Rafael and Sell{\'e}s-Navarro, Inmaculada and Palaz{\'o}n-Cabanes, Ana and Sancho-G{\'o}mez, Jos{\'e}-Luis},
  journal={Scientific Data},
  volume={9},
  number={1},
  pages={291},
  year={2022},
  publisher={Nature Publishing Group}
}

@article{zhang2023coph100,
  title={{COph100}: A Comprehensive Ophthalmic Fundus Image Dataset for Deep Learning},
  author={Zhang, Xiaoming and others},
  journal={Scientific Data},
  year={2023},
  note={Fundus image dataset for retinal vessel segmentation}
}

@article{hatamizadeh2022ravir,
  title={{RAVIR}: A dataset and methodology for the semantic segmentation of retinal arteries and veins in infrared reflectance imaging},
  author={Hatamizadeh, Ali and Hosseini, Hamid and Patel, Niraj and Choi, Jinsoo and Pole, Cameron C and Hoeferlin, Cory M and Schwartz, Steven D and Terzopoulos, Demetri},
  journal={IEEE Journal of Biomedical and Health Informatics},
  volume={26},
  number={7},
  pages={3272--3283},
  year={2022},
  publisher={IEEE}
}

@article{staal2004drive,
  title={Ridge-based vessel segmentation in color images of the retina},
  author={Staal, Joes and Abramoff, Michael D and Niemeijer, Meindert and Viergever, Max A and Van Ginneken, Bram},
  journal={IEEE Transactions on Medical Imaging},
  volume={23},
  number={4},
  pages={501--509},
  year={2004},
  publisher={IEEE}
}

@article{kashani2021intraretinal,
  title={Automated Intraretinal Cystoid Fluid Segmentation Using Optical Coherence Tomography Images and Deep Learning},
  author={Kashani, Amir H and others},
  journal={Translational Vision Science \& Technology},
  volume={10},
  number={14},
  pages={23},
  year={2021},
  publisher={ARVO}
}

@article{codella2019isic,
  title={Skin Lesion Analysis Toward Melanoma Detection 2018: A Challenge Hosted by the {International Skin Imaging Collaboration (ISIC)}},
  author={Codella, Noel and Rotemberg, Veronica and Tschandl, Philipp and Celebi, M Emre and Dusza, Stephen and Gutman, David and Helba, Brian and Kalloo, Aadi and Liopyris, Konstantinos and Marchetti, Michael and others},
  journal={arXiv preprint arXiv:1902.03368},
  year={2019}
}

@article{moreira2012mmdb,
  title={{INbreast}: Toward a Full-field Digital Mammographic Database},
  author={Moreira, In{\^e}s C and Amaral, Igor and Domingues, In{\^e}s and Cardoso, Ant{\'o}nio and Cardoso, Maria Jo{\~a}o and Cardoso, Jaime S},
  journal={Academic Radiology},
  volume={19},
  number={2},
  pages={236--248},
  year={2012},
  publisher={Elsevier}
}

@article{bano2020fetoplac,
  title={{FetReg}: Fetoscopic Placental Vessel Segmentation and Registration},
  author={Bano, Sophia and Casella, Alessandro and Vasconcelos, Francisco and Qayyum, Abdul and others},
  journal={Medical Image Analysis},
  volume={76},
  pages={102330},
  year={2022},
  publisher={Elsevier}
}

@inproceedings{long2015fcn,
  title={Fully convolutional networks for semantic segmentation},
  author={Long, Jonathan and Shelhamer, Evan and Darrell, Trevor},
  booktitle=CVPR,
  pages={3431--3440},
  year={2015}
}

@inproceedings{ronneberger2015unet,
  title={U-Net: Convolutional networks for biomedical image segmentation},
  author={Ronneberger, Olaf and Fischer, Philipp and Brox, Thomas},
  booktitle={MICCAI},
  pages={234--241},
  year={2015},
  organization={Springer}
}

@article{oktay2018attentionunet,
  title={Attention U-Net: Learning Where to Look for the Pancreas},
  author={Oktay, Ozan and Schlemper, Jo and Folgoc, Loic Le and Lee, Matthew and Heinrich, Mattias and Misawa, Kazunari and Mori, Kensaku and McDonagh, Steven and Hammerla, Nils Y and Kainz, Bernhard and others},
  journal={arXiv preprint arXiv:1804.03999},
  year={2018}
}

@inproceedings{zhou2018unetpp,
  title={UNet++: A Nested U-Net Architecture for Medical Image Segmentation},
  author={Zhou, Zongwei and Rahman Siddiquee, Md Mahfuzur and Tajbakhsh, Nima and Liang, Jianming},
  booktitle={MICCAI},
  pages={3--11},
  year={2018},
  organization={Springer}
}

@inproceedings{cicek2016unet3d,
  title={3D U-Net: learning dense volumetric segmentation from sparse annotation},
  author={{\c{C}}i{\c{c}}ek, {\"O}zg{\"u}n and Abdulkadir, Ahmed and Lienkamp, Soeren and Brox, Thomas and Ronneberger, Olaf},
  booktitle={MICCAI},
  pages={424--432},
  year={2016},
  organization={Springer}
}

@inproceedings{milletari2016vnet,
  title={V-Net: Fully Convolutional Neural Networks for Volumetric Medical Image Segmentation},
  author={Milletari, Fausto and Navab, Nassir and Ahmadi, Seyed-Ahmad},
  booktitle={3DV},
  pages={565--571},
  year={2016},
  organization={IEEE}
}

@article{isensee2021nnunet,
  title={nnU-Net: a self-configuring method for deep learning-based biomedical image segmentation},
  author={Isensee, Fabian and Jaeger, Paul F and Kohl, Simon AA and Petersen, Jens and Maier-Hein, Klaus H},
  journal={Nature Methods},
  volume={18},
  number={2},
  pages={203--211},
  year={2021},
  publisher={Nature Publishing Group}
}

@article{wasserthal2023totalsegmentator,
  title={TotalSegmentator: Robust Segmentation of 104 Anatomic Structures in CT Images},
  author={Wasserthal, Jakob and Breit, Hanns-Christian and Meyer, Manfred T and Pradella, Maurice and Hinck, Daniel and Sauter, Alexander W and Heye, Tobias and Boll, Daniel T and Cyriac, Joshy and Yang, Shan and others},
  journal={Radiology: Artificial Intelligence},
  volume={5},
  number={5},
  year={2023},
  publisher={RSNA}
}

@article{chen2021transunet,
  title={TransUNet: Transformers Make Strong Encoders for Medical Image Segmentation},
  author={Chen, Jieneng and Lu, Yongyi and Yu, Qihang and Luo, Xiangde and Adeli, Ehsan and Wang, Yan and Lu, Le and Yuille, Alan L and Zhou, Yuyin},
  journal={arXiv preprint arXiv:2102.04306},
  year={2021}
}

@inproceedings{hatamizadeh2022unetr,
  title={UNETR: Transformers for 3D Medical Image Segmentation},
  author={Hatamizadeh, Ali and Tang, Yucheng and Nath, Vishwesh and Yang, Dong and Myronenko, Andriy and Landman, Bennett and Roth, Holger R and Xu, Daguang},
  booktitle={WACV},
  pages={574--584},
  year={2022},
  organization={IEEE}
}

@inproceedings{hatamizadeh2022swinunetr,
  title={Swin UNETR: Swin Transformers for Semantic Segmentation of Brain Tumors in MRI Images},
  author={Hatamizadeh, Ali and Nath, Vishwesh and Tang, Yucheng and Yang, Dong and Roth, Holger R and Xu, Daguang},
  booktitle={MICCAI},
  year={2022},
  organization={Springer}
}

@article{zhou2021nnformer,
  title={nnFormer: Interleaved Transformer for Volumetric Segmentation},
  author={Zhou, Hong-Yu and Guo, Jiansen and Zhang, Yinghao and Yu, Lequan and Wang, Liansheng and Yu, Yizhou},
  journal={arXiv preprint arXiv:2109.03201},
  year={2021}
}

@article{gu2023mamba,
  title={Mamba: Linear-Time Sequence Modeling with Selective State Spaces},
  author={Gu, Albert and Dao, Tri},
  journal={arXiv preprint arXiv:2312.00752},
  year={2023}
}

@inproceedings{ma2024umamba,
  title={U-Mamba: Enhancing Long-range Dependency for Biomedical Image Segmentation},
  author={Ma, Jun and Li, Feifei and Wang, Bo},
  booktitle={MICCAI},
  year={2024}
}

@inproceedings{xing2024segmamba,
  title={SegMamba: Long-range Sequential Modeling Mamba For 3D Medical Image Segmentation},
  author={Xing, Zhaohu and Ye, Tian and Yang, Yijun and Liu, Guang and Zhu, Lei},
  booktitle={MICCAI},
  year={2024}
}

@article{ruan2024vmunet,
  title={VM-UNet: Vision Mamba UNet for Medical Image Segmentation},
  author={Ruan, Jiacheng and Xiang, Suncheng},
  journal={arXiv preprint arXiv:2402.02491},
  year={2024}
}

@inproceedings{Liu_SwinUMamba_MICCAI2024,
  title={Swin-UMamba: Mamba-based UNet with ImageNet-based pretraining},
  author={Liu, Jiarun and Yang, Hao and Caverly, Hong-Yu and others},
  booktitle={MICCAI},
  year={2024}
}

@inproceedings{lueddecke2022clipseg,
  title={Image Segmentation Using Text and Image Prompts},
  author={L{\"u}ddecke, Timo and Ecker, Alexander},
  booktitle=CVPR,
  pages={7086--7096},
  year={2022}
}

@inproceedings{rao2022denseclip,
  title={DenseCLIP: Language-Guided Dense Prediction with Context-Aware Prompting},
  author={Rao, Yongming and Zhao, Wenliang and Liu, Guangyi and Lu, Jiwen and Zhou, Jie},
  booktitle=CVPR,
  pages={18082--18091},
  year={2022}
}

@inproceedings{ding2023maskclip,
  title={MaskCLIP: Mask Transformer for Open-Vocabulary Universal Image Segmentation},
  author={Ding, Zheng and Wang, J. and Tu, Zhuowen},
  booktitle=CVPR,
  year={2023}
}

@inproceedings{liang2023maskadaptedclip,
  title={Open-Vocabulary Semantic Segmentation with Mask-adapted CLIP},
  author={Liang, Feng and Wu, Bichen and Dai, Xiaoliang and Li, Kunpeng and Zhao, Yinan and Zhang, Hang and Zhang, Peizhao and Vajda, Peter and Marculescu, Diana},
  booktitle=CVPR,
  year={2023}
}

@inproceedings{wang2022cris,
  title={CRIS: CLIP-Driven Referring Image Segmentation},
  author={Wang, Zhaoqing and Lu, Yu and Li, Qiang and Tao, Xiangtai and Guo, Yandong and Gong, Mingming and Liu, Tongliang},
  booktitle=CVPR,
  pages={11686--11695},
  year={2022}
}

@inproceedings{yang2022lavt,
  title={LAVT: Language-Aware Vision Transformer for Referring Image Segmentation},
  author={Yang, Zhao and Wang, Jiaqi and Tang, Yansong and Chen, Kai and Zhao, Hengshuang and Torr, Philip HS},
  booktitle=CVPR,
  pages={18155--18165},
  year={2022}
}

@article{wang2018deepigeos,
  title={DeepIGeoS: A Deep Interactive Geodesic Framework for Medical Image Segmentation},
  author={Wang, Guotai and Zuluaga, Maria A and Li, Wenqi and Pratt, Rosalind and Patel, Premal A and Aertsen, Michael and Doel, Tom and David, Anna L and Deprest, Jan and Ourselin, Sebastien and Vercauteren, Tom},
  journal=PAMI,
  volume={41},
  number={7},
  pages={1559--1572},
  year={2018},
  publisher={IEEE}
}

@inproceedings{kirillov2023sam,
  title={Segment Anything},
  author={Kirillov, Alexander and Mintun, Eric and Ravi, Nikhila and Mao, Hanzi and Rolland, Chloe and Gustafson, Laura and Xiao, Tete and Whitehead, Spencer and Berg, Alexander C and Lo, Wan-Yen and Doll{\'a}r, Piotr and Girshick, Ross},
  booktitle=ICCV,
  pages={4015--4026},
  year={2023}
}

@article{ravi2024sam2,
  title={SAM 2: Segment Anything in Images and Videos},
  author={Ravi, Nikhila and Gabeur, Valentin and Hu, Yuan-Ting and Hu, Ronghang and Ryali, Chaitanya and Ma, Tengyu and Khedr, Haitham and R{\"a}dle, Roman and Rolland, Chloe and Gustafson, Laura and others},
  journal={arXiv preprint arXiv:2408.00714},
  year={2024}
}

@article{ma2024medsam,
  title={MedSAM: Segment Anything in Medical Images},
  author={Ma, Jun and He, Yuting and Li, Feifei and Han, Lin and You, Chuyang and Wang, Bo},
  journal={Nature Communications},
  volume={15},
  pages={654},
  year={2024},
  publisher={Nature Publishing Group}
}

@article{wu2025medsa,
  title={Medical SAM Adapter: Adapting Segment Anything Model for Medical Image Segmentation},
  author={Wu, Junde and Fu, Rao and Fang, Huihui and Liu, Yuanpei and Wang, Zhaowei and Xu, Yanwu and Jin, Yueming and Arbel, Tal},
  journal={Medical Image Analysis},
  volume={102},
  pages={103547},
  year={2025},
  publisher={Elsevier}
}

@article{wang2023sammed3d,
  title={SAM-Med3D},
  author={Wang, Haoyu and Guo, Sizheng and Ye, Jin and Deng, Zhongying and Ren, Yanzhou and Li, Yida and Wan, Xiang},
  journal={arXiv preprint arXiv:2310.15161},
  year={2023}
}

@article{ma2025medsam2,
  title={Medical SAM 2: Segment medical images as video via Segment Anything Model 2},
  author={Ma, Jun and Zhu, Yutong and Wang, Bo},
  journal={arXiv preprint arXiv:2408.00874},
  year={2025},
  note={Updated version of MedSAM-2}
}

@inproceedings{ye2023uniseg,
  title={UniSeg: A Prompt-driven Universal Segmentation Model as well as A Strong Representation Learner},
  author={Ye, Yiwen and Xie, Yutong and Zhang, Jianpeng and Chen, Ziyang and Xia, Yong},
  booktitle={MICCAI},
  pages={508--518},
  year={2023},
  organization={Springer}
}

@article{ye2024meduniseg,
  title={MedUniSeg: 2D and 3D Medical Image Segmentation via a Prompt-driven Universal Model},
  author={Ye, Yiwen and Chen, Ziyang and Zhang, Jianpeng and Xie, Yutong and Xia, Yong},
  journal={arXiv preprint arXiv:2410.05905},
  year={2024}
}

@inproceedings{wu2024oneprompt,
  title={One-Prompt to Segment All Medical Images},
  author={Wu, Junde and Min, Xu},
  booktitle=CVPR,
  year={2024}
}

@article{carion2025sam3,
  title={SAM 3: Segment Anything with Concepts},
  author={Carion, Nicolas and others},
  journal={arXiv preprint arXiv:2511.16719},
  year={2025}
}

@article{zhao2025sat,
  title={One Model to Rule them All: Towards Universal Segmentation for Medical Images with Text Prompts},
  author={Zhao, Ziheng and others},
  journal={arXiv preprint arXiv:2312.17183},
  year={2025},
  note={Also referred to as SAT}
}

@article{litjens2017survey,
  title={A survey on deep learning in medical image analysis},
  author={Litjens, Geert and Kooi, Thijs and Bejnordi, Babak Ehteshami and Setio, Arnaud Arindra Adiyoso and Ciompi, Francesco and Ghafoorian, Mohsen and Van Der Laak, Jeroen AWM and Van Ginneken, Bram and S{\'a}nchez, Clara I},
  journal={Medical Image Analysis},
  volume={42},
  pages={60--88},
  year={2017},
  publisher={Elsevier}
}

@article{esteva2017dermatologist,
  title={Dermatologist-level classification of skin cancer with deep neural networks},
  author={Esteva, Andre and Kuprel, Brett and Novoa, Roberto A and Ko, Justin and Swetter, Susan M and Blau, Helen M and Thrun, Sebastian},
  journal={Nature},
  volume={542},
  number={7639},
  pages={115--118},
  year={2017},
  publisher={Nature Publishing Group}
}

@article{shen2017deep,
  title={Deep learning in medical image analysis},
  author={Shen, Dinggang and Wu, Guorong and Suk, Heung-Il},
  journal={Annual Review of Biomedical Engineering},
  volume={19},
  pages={221--248},
  year={2017},
  publisher={Annual Reviews}
}

@article{guan2021domain,
  title={Domain adaptation for medical image analysis: a survey},
  author={Guan, Hao and Liu, Mingxia},
  journal={IEEE Transactions on Biomedical Engineering},
  volume={69},
  number={3},
  pages={1173--1185},
  year={2021},
  publisher={IEEE}
}

@inproceedings{liu2021feddg,
  title={FedDG: Federated Domain Generalization on Medical Image Segmentation via Episodic Learning in Continuous Frequency Space},
  author={Liu, Quande and Chen, Cheng and Qin, Jing and Dou, Qi and Heng, Pheng-Ann},
  booktitle={CVPR},
  pages={1013--1023},
  year={2021}
}

@article{he2024accuracy,
  title={Accuracy of Segment-Anything Model (SAM) in medical image segmentation: a comprehensive evaluation},
  author={He, Sheng and Bao, Rina and Grant, P Ellen and Ou, Yangming},
  journal={International Journal of Computer Assisted Radiology and Surgery},
  volume={19},
  number={1},
  pages={31--46},
  year={2024},
  publisher={Springer}
}

@article{mazurowski2023segment,
  title={Segment anything model for medical image analysis: an experimental study},
  author={Mazurowski, Maciej A and Dong, Haoyu and Gu, Hanxue and Yang, Jichen and Konz, Nicholas and Zhang, Yixin},
  journal={Medical Image Analysis},
  volume={89},
  pages={102918},
  year={2023},
  publisher={Elsevier}
}

@article{huang2024segment,
  title={Segment anything model for medical images?},
  author={Huang, Yuhao and Yang, Xin and Liu, Liu and Zhou, Han and Chang, Ao and Zhou, Xinrui and Chen, Rusi and Yu, Junxuan and Chen, Jossi and Chen, Chaanyu and others},
  journal={Medical Image Analysis},
  volume={92},
  pages={103061},
  year={2024},
  publisher={Elsevier}
}

@article{shao2025icm,
  title={ICM-Fusion: In-Context Meta-Optimized LoRA Fusion for Multi-Task Adaptation},
  author={Shao, Yihua and Lin, Xiaofeng and Long, Xinwei and Chen, Siyu and Yan, Minxi and Liu, Yang and Yan, Ziyang and Ma, Ao and Tang, Hao and Guo, Jingcai},
  journal={arXiv preprint arXiv:2508.04153},
  year={2025}
}

@article{shao2024gwq,
  title={GWQ: Gradient-Aware Weight Quantization for Large Language Models},
  author={Shao, Yihua and Liang, Siyu and Ling, Zijian and Yan, Minxi and Liu, Haiyang and Chen, Siyu and Yan, Ziyang and Zhang, Chenyu and Qin, Haotong and Magno, Michele and others},
  journal={arXiv preprint arXiv:2411.00850},
  year={2024}
}

@article{shao2025context,
  title={In-Context Meta LoRA Generation},
  author={Shao, Yihua and Yan, Minxi and Liu, Yang and Chen, Siyu and Chen, Wenjie and Long, Xinwei and Yan, Ziyang and Li, Lei and Zhang, Chenyu and Sebe, Nicu and others},
  journal={arXiv preprint arXiv:2501.17635},
  year={2025}
}

@article{shao2025tr,
  title={TR-DQ: Time-Rotation Diffusion Quantization},
  author={Shao, Yihua and Lin, Deyang and Zeng, Fanhu and Yan, Minxi and Zhang, Muyang and Chen, Siyu and Fan, Yuxuan and Yan, Ziyang and Wang, Haozhe and Guo, Jingcai and others},
  journal={arXiv preprint arXiv:2503.06564},
  year={2025}
}

@article{yan20243dsceneeditor,
  title={3dsceneeditor: Controllable 3d scene editing with gaussian splatting},
  author={Yan, Ziyang and Li, Lei and Shao, Yihua and Chen, Siyu and Wu, Zongkai and Hwang, Jenq-Neng and Zhao, Hao and Remondino, Fabio},
  journal={arXiv preprint arXiv:2412.01583},
  year={2024}
}

@inproceedings{yan2025renderworld,
  title={Renderworld: World model with self-supervised 3d label},
  author={Yan, Ziyang and Dong, Wenzhen and Shao, Yihua and Lu, Yuhang and Liu, Haiyang and Liu, Jingwen and Wang, Haozhe and Wang, Zhe and Wang, Yan and Remondino, Fabio and others},
  booktitle={2025 IEEE International Conference on Robotics and Automation (ICRA)},
  pages={6063--6070},
  year={2025},
  organization={IEEE}
}

@inproceedings{shao2025eventvad,
  title={Eventvad: Training-free event-aware video anomaly detection},
  author={Shao, Yihua and He, Haojin and Li, Sijie and Chen, Siyu and Long, Xinwei and Zeng, Fanhu and Fan, Yuxuan and Zhang, Muyang and Yan, Ziyang and Ma, Ao and others},
  booktitle={Proceedings of the 33rd ACM International Conference on Multimedia},
  pages={2586--2595},
  year={2025}
}

@inproceedings{shao2025accidentblip,
  title={Accidentblip: Agent of accident warning based on ma-former},
  author={Shao, Yihua and Xu, Yeling and Long, Xinwei and Chen, Siyu and Yan, Ziyang and Liu, Haoting and Wang, Yan and Tang, Hao and Yang, Yang},
  booktitle={2025 IEEE Intelligent Vehicles Symposium (IV)},
  pages={2156--2161},
  year={2025},
  organization={IEEE}
}
\nocite{*}


\end{document}